\documentclass[11pt,a4paper]{article}
\usepackage[acceptedWithA]{tacl2021v1}
\usepackage{times}
\usepackage{latexsym}
\usepackage[T1]{fontenc}
\usepackage[utf8]{inputenc}
\usepackage{microtype}
\usepackage{inconsolata}
\usepackage{graphicx}
\usepackage{multirow}
\usepackage{amsmath,amssymb,amsfonts}
\usepackage{amsthm}
\usepackage{mathrsfs}
\usepackage[title]{appendix}
\usepackage{xcolor}
\usepackage{textcomp}
\usepackage{manyfoot}
\usepackage{booktabs}
\usepackage{algorithm}
\usepackage{algorithmicx}
\usepackage{algpseudocode}
\usepackage{listings}
\usepackage{comment}
\usepackage{booktabs}
\usepackage{multirow}
\usepackage{tikz}
\def\checkmark{\tikz\fill[scale=0.4](0,.35) -- (.25,0) -- (1,.7) -- (.25,.15) -- cycle;} 

\def\crosstick{
  \tikz\draw[scale=0.4, line width=0.6pt]
    (0.2,0.2) -- (0.8,0.8)
    (0.2,0.8) -- (0.8,0.2);
}

\title{\textit{Versteasch du mi?} Computational and Socio-Linguistic Perspectives on GenAI, LLMs, and Non-Standard Language}

\author{
    \begin{tabular}{c@{\hspace{1.5em}}c}
    { Verena Platzgummer \qquad\qquad John McCrae} &  Sina Ahmadi \\
    {\normalfont University of Galway} & {\normalfont University of Zurich} \\
    {\normalfont \texttt{\{verena.platzgummer,john.mccrae\}@universityofgalway.ie}} &
    {\normalfont \texttt{sina.ahmadi@uzh.ch}}
    \end{tabular}
}

\begin{document}
\maketitle
\thispagestyle{plain}
\pagestyle{plain}

\begin{abstract}
    The design of Large Language Models (LLMs) and generative artificial intelligence (GenAI) has been shown to be ``unfair'' to less-spoken languages~\citep{petrov2023language} and to deepen the digital language divide ~\citep{bella2023towards}. Critical sociolinguistic work has also argued that these technologies are not only made possible by prior socio-historical processes of linguistic standardisation, often grounded in European nationalist and colonial projects ~\citep{migge2025material}, but also exacerbate epistemologies of language as ``monolithic, monolingual, syntactically standardized systems of meaning'' \cite[p.~5]{schneider2024sociolinguist}. In our paper, we draw on earlier work on the intersections of technology and language policy ~\citep{kelly2019multilingualism} and bring our respective expertise in critical sociolinguistics and computational linguistics to bear on an interrogation of these arguments. We take two different complexes of non-standard linguistic varieties in our respective repertoires--\textbf{South Tyrolean} dialects, which are widely used in informal communication in South Tyrol, Italy ~\citep{alber2024verschriftungsprinzipien}, as well as varieties of \textbf{Kurdish}--as starting points to an interdisciplinary exploration of the intersections between GenAI and linguistic variation and standardisation. We discuss both how LLMs can be made to deal with non-standard language from a technical perspective, and whether, when or how this can contribute to ``democratic and decolonial digital and machine learning strategies'' \cite[p.~12]{migge2025material}, which has direct policy implications.
\end{abstract}

\section{Introduction}

The rapid advancement of Large Language Models (LLMs) and Generative Artificial Intelligence (GenAI) has transformed digital communication, yet these technologies systematically privilege standardized and, in computational linguistic terms, high-resource languages while marginalizing millions of speakers who communicate primarily through non-standard varieties and dialects. Recent scholarship demonstrates that LLM design is fundamentally ``unfair'' to less-spoken languages~\citep{petrov2023language} and deepens the digital language divide~\citep{bella2023towards, birnie2026no}. Critical sociolinguistic work argues that these technologies not only emerge from historical processes of linguistic standardization rooted in colonial and nationalist projects~\citep{migge2025material}, but also reinforce epistemologies of language as ``monolithic, monolingual, syntactically standardized systems of meaning''~\cite[p.~5]{schneider2024sociolinguist}. When AI systems fail to process non-standard varieties, including historically minoritized languages, they exclude speakers from full access to digital spaces~\citep{birnie2026no}. Within computational linguistics and Natural Language Processing (NLP) communities, there have also been calls for decolonial~\citep{bird-2020-decolonising} or speaker-centric ~\citep{ramponi-2024-language} approaches to language technologies that place the needs of speech communities rather than computational advances on center stage.

This paper brings together critical sociolinguistics and computational linguistics to critically examine systemic inequalities that speakers of non-standard languages and varieties face in relation to language technologies. We analyze the technical barriers preventing LLMs from processing non-standard varieties, examine how evaluation frameworks reproduce standardization biases, and explore policy implications for diverse actors—from Big Tech and states to civil society and academia. We support our discussion with two complementary case studies of non-standard varieties we are speakers of ourselves: South Tyrolean German (Verena) and Kurdish varieties (Sina). While both cases illustrate the limited support that non-standard varieties receive in contemporary language technologies, they differ significantly in the forces that shape their digital marginalization. South Tyrolean dialect comprises non-standard forms of German widely used in informal communication in the Italian province of South Tyrol~\citep{alber2024verschriftungsprinzipien}, both in speaking and writing. The variety belongs to the wider category of South Bavarian dialects, spoken in most of Austria and parts of Bavaria, and distinguishes itself from those mostly through effects of language contact with Italian as well as lexical particularities emerging from the different national context~\citep{leonardi2020hardly, abel2018bars}. It can be estimated that the complex of South Tyrolean German varieties is the primary medium of everyday interaction for about 300,000 of the province's inhabitants ~\citep{stat_report_key}, yet it has been largely absent from LLM training data. Its relative marginalization is therefore less a consequence of political exclusion than of limited visibility within computational infrastructures, where it is largely subsumed under broader linguistic categories and rarely evaluated in its own right. Kurdish, by contrast, represents a dialect continuum spoken by over 40 million people across multiple nation-states, characterized by orthographic diversity, systematic political suppression, and acute digital underrepresentation. While Central Kurdish (Sorani) and Northern Kurdish (Kurmanji) have achieved modest computational resources, varieties such as Southern Kurdish, Hawrami, and Zazaki remain almost entirely invisible to language technology. 

Building on earlier work examining technology and language policy intersections~\citep{kelly2019multilingualism}, we argue that addressing LLM marginalization of non-standard varieties requires more than technical fixes; it demands a reorientation toward ``democratic and decolonial digital and machine learning strategies''~\citep{migge2025material} that center linguistic agency and treat variation as constitutive of human communication rather than noise to eliminate. This has direct policy implications, from requiring ``dialect gap'' reporting to ensuring community data sovereignty. Ultimately, whether LLMs should ``handle'' non-standard varieties is not primarily a technical question but a fundamentally political one, with implications for digital sovereignty, cultural preservation, and social justice. We argue that it is only by embracing variation that the world's linguistic diversity can truly enter digital spaces.

In the following sections, we will first address sociolinguistic (section~\ref{section_sociolinguistic}) and computational linguistic perspectives on standard and non-standard language and language technologies (section~\ref{section_cl_llm}), before giving an overview of the policy landscape in relation to linguistic variation and LLMs (section~\ref{section_llm_policy}). We will proceed by examining the cases of South Tyrolean dialects (section~\ref{case_study_tyrolean}) and varieties of Kurdish (section~\ref{case_study_kurdish}) and end with overarching conclusions and policy implications.

\section{Sociolinguistic Perspectives on (Non-)Standard Language and LLMs}
\label{section_sociolinguistic}
The concept of\textit{ standard language}, as well as processes of linguistic standardisation, have been objects of sociolinguistic investigation since the inception of the field itself ~\citep{haugen1959planning,ferguson1962language}. The impetus for this type of work largely came from concurrent processes of standardization in newly independent states, often postcolonial realities. As such, it is closely linked to the beginnings of language planning and policy as a subfield of applied linguistics. As characteristics of standard languages,~\citet{auer2005europe} lists their validity across regional borders, their being considered as the ‘best’ language within their realm of validity, and their being codified in norms. As~\citet{deumert2004language,deumert2010imbodela} notes, it is their institutional enforcement that distinguishes standard languages from other types of linguistic norms, both from “always emergent, variable, and never ‘fixed’ conventions”~\citep[p. 244]{deumert2010imbodela} and from language standards that go beyond such conventions in that they become morally imperative, but are not quite standard languages as they are not institutionally enforced. 

Processes of standardisation usually involve some degree of reduction in linguistic variability~\citep{deumert2004language,milroy2012authority}. As such, these processes can also be understood as linguistic hierarchisations~\citep{kristiansen2011standard,costa2017standardising,gal2017visions}, whereby specific sets of linguistic features, as well as entire varieties conceived as bundles of such features, are imbued with legitimacy for public use, while others are considered acceptable only in private~\citep{costa2017standardising}. As~\citet{gal2017visions}  argues, this legitimacy and authority of linguistic forms stem from the two ideological complexes of anonymity and authenticity that go back to rationalist and romantic philosophies of the 17th and 18th centuries.

It is commonly recognised that “[s]tandardization is […] best approached as an ideological phenomenon”~\citep[p. 222]{gal2017visions}. Processes of linguistic standardisation have tended to be part and parcel of processes of nation-building~\citep{deumert2010imbodela,costa2017standardising,erdocia2025language} and the concomitant creation of an apparently neutral public sphere, as well as of colonialism, and are quintessentially modernist projects~\citep{costa2017standardising}. Standard language and non-standard language – be it referred to as dialect, patois, etc. – are constructed in opposition to one another, whereby only standard language indexes modernist values like progress and development, and is associated with the future~\citep{gal2017visions}. It is easy to see how, within such standardisation regimes, the majority of the world’s languages that do not have standardised forms~\citep{Romaine2008} can become constructed as ‘lacking’ and ‘backwards’~\citep{gal2017visions}, and even more so the myriad of primarily oral language practices~\citep{bird-2020-decolonising}. In fact, the epistemologies of language underlying standard language ideologies are those of language “as an autonomous and unitary system whose main function is the effective and precise transmission of information”~\citep[p. 245]{deumert2010imbodela}, disregarding a view of language as situated and embodied social practice~\citep{costa2017standardising,schneider2024sociolinguist}.

It has been argued that the development of LLMs has been based both on the epistemological notions underlying standardising regimes and on the effects of these regimes on language practices: According to~\citet{schneider2024sociolinguist}, LLMs ''build on a foundation of prior technological and sociopolitical conditions including phonetic spelling, standardized print literacy, and linguistic nationalism``. It is thus the diffusion of notions of standard language along with universal education and mass literacy that have produced a quantity of text homogeneous enough to be probabilistically modellable – and these models, in turn, provide the basis for the generation of statistically likely sequences of tokens by generative AI~\citep{schneider2024sociolinguist}. 

While linguistic standardisation is usually examined within the bounds of a single nation, a different perspective is necessary for investigating the development and effects of language technologies like LLMs and generative AI. As~\citet[p. 381]{schneider2022multilingualism} notes, in contrast with the modernist projects of nation-building and linguistic standardisation, these technologies do not aim “to homogenize language in order to create a linguistically homogenous national population” but instead produce a global digital public and are motivated, in most cases, by commercial interests. Previous work on the intersections of technology and language policy in relation to the development of the internet has shown how “technological advances and breakthroughs occur in particular ideological and cultural spaces, and the shape of those technological advances bears the imprint of those cultural and ideological norms”~\citep[p. 27]{kelly2019multilingualism}. Dividing the de facto language policy of the internet into four distinct periods,~\citet{kelly2019multilingualism} argued that we are currently witnessing the period of idiolingualism, characterised by mass linguistic customization. It is to be expected that this type of customization or personalization will impact on the linguistic direction that LLMs will take, and that user needs – as well as their power of consumption – will contribute to steering their development~\citep{kelly2025artificial,schneider2022multilingualism}

In this manner, language technologies may reinforce and potentially also reconfigure linguistic hierarchisations~\citep{schneider2022multilingualism,leblebici2025alexa}. As we will also show in our next section, the hierarchies originating from the interplay between linguistic standardisation and colonialism are already being exacerbated, with LLMs further contributing to the dominance of English and of other European-derived standard languages~\citep{schneider2022multilingualism}, and of Mandarin~\citep{arora2019next}. At the same time, however, the jury is still out on the effects that this technology will have on variation within what is commonly constructed as one language. For instance, the fact that the acceptance of standard languages as ‘best’ language tends to be more wide-spread than their use~\citep{kristiansen2011standard,gal2017visions} might mean that it might not be those forms of language that LLMs will reinforce, if other forms end up being used more frequently – analogous to destandardisation tendencies that have been identified in different European contexts for some time now~\citep{auer2005europe}. It thus becomes a highly relevant question how LLMs and generative AI will impact on hierarchisations of language practices and their associated types of speakers. 

% https://aclanthology.org/2025.acl-long.370.pdf
\section{Computational Linguistic Perspectives on LLMs \& (Non-)Standard Language}
\label{section_cl_llm}
Underlying those generative AI systems that generate text, LLMs have demonstrated remarkable capabilities in what computational linguistics refers to as high-resource linguistic environments~\citep{humanitys_last_exam}, especially English. However, their deployment across the global linguistic landscape remains deeply asymmetrical. In computational linguistic terms, the ``digital divide'' between languages has translated into lesser-used languages in the digital sphere becoming classed as \textit{Under-Resourced Languages} (URLs), lacking representative training data and thus \textit{language resources}. At the same time, these languages are impaired by Anglo-centric architectural biases of LLMs, which negatively affect performance~\citep{globalmmlu}. It has been shown that models will produce responses that are not valid in the cultural context of the language, thus reducing diversity and leading to stereotyping~\citep{veselovsky-etal-2026-localized}. While this challenge is substantial for regional minority languages, like Basque, or even lesser-used national languages such as Irish or Estonian, non-standardized dialects and varieties have been barely considered when evaluating factors such as LLM performance, even though LLMs can work to some degree in such languages~\citep{faisal-etal-2025-dialectal}. Applying LLMs to these languages is often complex due to issues such as the lack of formal orthographies, diglossic tension with standard versions (like Standard German~\citep{lin-etal-2025-construction}) and they are even frequently deliberately excluded from the massive web-scrapes that form LLM training sets~\citep{the_pile}. Consequently, these languages face a double marginalization where both the data scarcity and the structural assumptions of modern NLP fail to capture their unique phonetic, syntactic, and cultural nuances. We see three main areas where current research on LLMs can be improved for non-standard languages: The architecture of LLMs, especially around the tokenization of the input text, the handling of morphological complexity and orthographic variation, and the creation of relevant benchmarks.

The technical infrastructure of language technologies significantly impacts dialects and determines whether a variety is even visible to digital tools. One issue in this regard is the assignment of ISO-639 codes\footnote{We refer primarily to the ISO-639-3 codes, which consist of three letters and cover many more languages than ISO 639-1 and ISO 639-2. Many primarily oral languages have only an ISO 639-3 code, such as Shabaki (\texttt{sdb}) or Luri Bakhtiari (\texttt{bqi}).}, which acts as a primary gatekeeper; for instance, German dialects like Upper Saxon or Bavarian have assigned ISO codes, which allows them to be catalogued in major Natural Language Processing (NLP) resources like OPUS and the Virtual Language Observatory\footnote{We note that some language varieties are encoded with the help of a regional code, e.g., \texttt{fr-CA} for \textit{Qu\'eb\'ecois}. However, this is limited as regional codes are not specific to regions within a country and this creates confusion with codes for orthographic variations, such as British and American English.}. Without such codes, whose colonial history~\citet{migge2025material} have elaborated on, a variety is effectively invisible to language technology, making it nearly impossible to track its representation or performance in AI models, although the assignment of an ISO code does not necessarily translate into meaningful support in the technology industry.
Moreover, the transcription of some dialects may make use of diacritics or other writing methods not supported by the Unicode standard, and as such cannot be processed. The graphemic representation of any language must therefore either remain within the existing coding, or the respective signs must be added to the Unicode standard ~\citep{gorman2025don, birnie2026no}. 

To see why LLMs struggle with non-standard languages, we must also look at the representation system that stands between human text and the machine, namely the \textit{tokenizer}. Modern AI models do not \emph{read} text as humans do, but instead convert the input text into a sequence of numbers that can be processed by a neural network. The process of segmenting the text is called tokenization and most current LLMs use methods based on Byte Pair Encoding~\citep[BPE,][]{bpe}. Instead of breaking text into whole words (which would create a vocabulary too large for the computer to manage) or individual characters (which are too small to carry much meaning), BPE identifies ``subword'' units based on how frequently they appear in a training dataset. As such models break words based on their statistical frequency, they will have more tokens assigned to languages that occur more frequently in the corpus. In this way, words for well-resourced languages will often be broken into fewer tokens and in ways that are more meaningful and related to the morphology of the language. For example, the English word `tokenization' is divided into two tokens by the widely used BERT model\footnote{\href{https://huggingface.co/google-bert/bert-base-uncased}{bert-based-uncased}}~\citep{bert}, namely `token+ization', separating the root and the suffix in a morphologically sound way. In contrast, the Irish word `ionchomharthú' (meaning `tokenization') is divided into 5 tokens, `ion+cho+m+hart+hú', and this tokenization not only does not bear any resemblance to the word's morphological components (ion+chomh+arth+ú), but even breaks the word across digraphs such as `mh' and `th'\footnote{Similar results hold for tokenizer in more recent LLMs. OpenAI provide a platform for this at \href{https://platform.openai.com/tokenizer}{https://platform.openai.com/tokenizer}}.

When a tokenizer encounters an under-resourced language or non-standard language, its training hasn't resulted in any specialized subword entries for that language. Instead, it produces a sequence of tokens formed from the subwords that it already has in its vocabulary. This is computationally inefficient, as a single dialectal word might be shattered into 5 or 6 tiny fragments (e.g., individual characters or meaningless byte-strings), whereas the English equivalent would be a single token. This computational inefficiency results in extra costs, e.g. for using more electricity to process a query in these languages. Providers often pass on this cost to users, charging them per token, which causes what is referred to as a "tokenization tax"; so speakers of such varieties literally pay more when they prompt generative AI in these varieties\footnote{\citet{petrov2023language} reports that most languages pay $2-3\times$ the price of English, with languages that use non-Latin scripts such as Arabic paying even higher costs up to $5\times$ and severely under-resourced languages such as Odia paying over $15\times$ the cost.} to process the same amount of information given in English. Performance, in terms of the model's ability, will also naturally be degraded as the model is working with small tokens without specific meaning, which carry less input than large subwords or whole words. Finally, most AI models have a context window~\citep{beltagy2020longformer}, which is a hard limit on how many tokens it can consider at one time. Because dialects require more tokens to express the same idea, they fill up the model's memory faster, leading to poorer reasoning and shorter possible conversations. It has been shown that language models are not able to robustly exploit long contexts~\citep{liu-etal-2024-lost}, and as dialects require more tokens per orthographic word, these issues are more likely to occur for such languages. In this light, the ``tokenization tax'' is more than a technical detail~\citep{ahia-etal-2023-languages} and instead represents algorithmic discrimination~\citep{benjamin2019race} that systematically increases the cost and decreases the quality of AI services for marginalized linguistic communities.

A second major issue that further compounds the challenge of tokenization is that non-standard languages have substantial orthographic variation and, like under-resourced languages more broadly, have high morphological complexity. 
%English and Chinese, languages spoken by many LLM developers, are 
English, as the dominant language of LLM development, is 
relatively ``morphologically poor'' (such as measured by \citet{morphological_richness}), having fewer morphemes and simpler morphotactics. Tokenization methods are inherently biased towards morphologically poor languages, as languages with a lower Type-To-Token (TTR) ratio~\citep{kettunen2014can} can be represented with a smaller vocabulary of subwords. 
This representational bias is particularly acute for non-standardised dialects; as \citet{kanjirangat-etal-2025-tokenization} demonstrates, %multilingual LLMs exhibit severe disparities in Tokenization Parity (TP) and Information Parity (IP), where claims of broad language support often mask deep mismatches at the token and script levels that degrade downstream performance. 
multilingual LLMs exhibit severe disparities in the number of tokens used, so that claims of broad language support do not match with often poor real-world performance for users of these languages.
Because tokenizers are optimised for the standardised, high-resource variants present in their pretraining corpora, they become brittle when encountering the unseen spelling variations inherent to closely related, non-standardised varieties~\citep{blaschke-etal-2023-manipulating}. This divergence frequently results in excessive subword splitting—where meaningful linguistic representations are fractured—rendering the split-word ratio a strong negative predictor of cross-lingual model performance~\citep{blaschke-etal-2023-manipulating}. To mitigate these surface-level mismatches without full architectural retraining, researchers have proposed strategies such as augmenting source-language training data with character-level noise to inherently make models robust against dialectal orthographic variation~\citep{aepli-sennrich-2022-improving}.

%On the other hand, Kurdish for example is a fusional language, where a single word can be built by combining morphemes for functions such as tense, person or negation. As such, a whole sentence in English may be compressed into a single complex Kurdish word. This challenge is intensified by the presence of allomorphs: Because Kurdish grammar is highly sensitive to the sounds surrounding a prefix or suffix, a single grammatical marker (like a plural or a tense indicator) might change its spelling or sound depending on the verb it attaches to. The development of language-specific tokenizers~\cite{kiulian-etal-2025-english} can be an effective remedy to this, however, it not only requires the development of a new tokenizer, but also a large amount of fine-tuning of the model to accommodate the new tokenizer. This is especially problematic in languages with a large number of code points, such as Chinese or Japanese, where tokenizers often break characters at the sub-character level, rendering tokens that have no linguistic connection to the input.

One specific challenge is morphologically complex languages, such as Kurdish, where a single word can be built by combining morphemes for functions such as tense, person or negation. As such, a whole sentence in English may be compressed into a single complex word. This challenge is intensified by the presence of allomorphs: languages such as Kurdish can be highly sensitive to the sounds surrounding a prefix or suffix, a single grammatical marker (like a plural or a tense indicator) might change its spelling or sound depending on the verb it attaches to. The development of language-specific tokenizers~\citep{kiulian-etal-2025-english} can be an effective remedy to this, however, it not only requires the development of a new tokenizer, but also a large amount of fine-tuning of the model to accommodate the new tokenizer.

For languages without a unified orthography, multiple spellings may exist based on local sub-dialects. Moreover, much of what LLMs see of such languages is extracted from informal contexts such as social media, which often contain errors due to carelessness (i.e., typos) as well as linguistic variation. Moreover, as these texts primarily come from informal contexts and specific domains, models' performance may not be as widely-applicable as for languages with a wider coverage.
Further, for non-standard varieties that are closely related to an existing standard language, there is a strong linguistic pull towards this standard (e.g. for South Tyrolean towards Modern High German) - not only for the speakers, who might draw on graphematic conventions of the associated standard language to spell their variety~\citep[e.g.][]{alber2024verschriftungsprinzipien}, but also for the model, which will have substantial training on the standard language. If AI tools are used for regional governance or service delivery, their inability to handle non-standard spelling can lead to exclusionary bias. Citizens who write in their native dialect may find themselves misunderstood or ignored by automated systems that were optimized for a standardized ``prestige'' language.

The final barrier to linguistic equity is how we measure LLM success. In AI development, ``what gets measured gets built''\footnote{Paraphrasing the famous quote by ``what gets measured gets managed'' by Peter Drucker}~\citep{what_gets_measured_gets_built}; however, the current tools for evaluating model performance are fundamentally ill-suited for non-standard and under-resourced languages. Measurement of an LLM’s general knowledge is achieved through benchmarks such as MMLU~\citep{mmlu}, which contains questions from textbooks and similar sources. The MMLU benchmark is highly US-centric, including topics such as US History, US Law and US Accounting, with approximately 28\% of the questions requiring specific knowledge of Western cultures and a staggering 84.9\% of geographic questions focus exclusively on North America or Europe~\citep{faisal-etal-2025-dialectal}. To address this, Global MMLU was introduced, which covers 42 languages, including highly under-resourced languages such as Nyanja or Telugu and explicitly marks questions as culturally sensitive. This avoids a distorted ranking, where a model might appear highly capable in a target language simply because it has memorised Western facts while failing to grasp the specific cultural, legal, or social nuances relevant to actual speakers of that language.

This bias is largely due to the fact that most benchmarks are created by translation of English benchmarks into other languages, and while this has been shown to correlate with human judgements~\citep{thellmann2024multilingualllmevaluationeuropean}, benchmarks are even worse for lower-resourced languages. This creates a translation pivot trap, where models are optimized to perform well on translated English concepts rather than achieving true native-level reasoning. To correct this bias, high-quality, culturally grounded benchmarks need to be developed, which is an immense financial and logistical undertaking. For example, the development of \textbf{MMLU-ProX}~\citep[covering 29 languages]{globalmmlu} involved a rigorous expert-review process to ensure cultural relevance, with development costs approaching \$80,000 at market rates. Furthermore, the quality of benchmarks created by translation varies significantly with STEM-related tasks exhibiting strong correlation with human judgements (0.70-0.85), while other tasks, such as question answering, have very poor correlation (0.11-0.30)~\citep{wu2025bitterlessonlearned2000}. These benchmarks, however, mainly reflect how closely a model approximates the language habits of the benchmark creator and may not reflect realistic usage patterns of speech communities.

The path forward requires moving away from translated benchmarks toward ``culturally and linguistically tailored benchmarks''~\citep{wu2025bitterlessonlearned2000}. A recent example is \textbf{IRLBench}~\citep{tran2025irlbenchmultimodalculturallygrounded}, a benchmark for the Irish language derived from the Irish Leaving Certificate exams.  Recently, DialectBench~\cite{dialectbench} has introduced a benchmark covering 281 dialects across 10 different tasks, providing the first benchmark that covers a wide variety of non-standard languages. Needless to say, this number is far from approaching a representation of the world's linguistic diversity. For instance, representation of the two case studies we address in this paper is only partial, in that the benchmark only includes wider Bavarian and not South Tyrolean, and only includes two of the five major Kurdish varieties. Further, there are specific aspects of relevance to non-standard languages that are excluded from benchmarks derived from English that are of relevance to speakers of the community. Firstly, being able to distinguish between dialects and measure whether a text is truly in the dialect or is just outputting standard language with a few dialectal words thrown in. This has led researchers to propose a Dialect Fidelity Score~\citep[DFS,][]{dialect_fidelity} to measure this. Secondly, more culturally relevant questions, e.g. explaining the moral of a specific proverb, would be relevant, where success requires an understanding of the cultural metaphors that do not exist in the model's English-centric training data. Finally, as non-standard languages are primarily used orally or in informal digital spaces, tasks in these benchmarks should reflect this bias. For example, VoxLect~\citep{voxlect} uses speech foundation models to evaluate how well AI understands regional accents and phonetic variations that are never captured in written text. Similarly, using ``noisy'' data from social media, such as in the NorDial benchmark for Norwegian dialects~\cite{barnes-etal-2021-nordial}, helps determining whether models can handle inconsistent orthography and non-standard spelling without crashing or defaulting to English or standard languages.

Considering the caveats of most current benchmarks, it is all the more alarming that even the best-performing models still show a persistent gap~\citep{ahuja-etal-2023-mega} and produce significantly worse results in languages other than English. For non-standard varieties, where formal academic benchmarks do not exist or are fragmentary, the ``performance cliff'' is likely even steeper. Without a policy-driven investment in local, non-translated evaluation data, speakers of non-standard varieties will not be able to make use of generative AI in these varieties.

\section{Linguistic Variation, Language Policy, and LLMs}
\label{section_llm_policy}
In this section, we turn to intersections between linguistic variation, language policy and LLMs and show how the way in which LLMs work is based on earlier language policy, and which types of policy actors respond in which kinds of ways to this technology and its development.

\subsection{LLM Functioning as Outcomes of Language Policy }

The functioning of today's LLMs can be seen in more than one way as outcomes of earlier language policies. For instance, efforts to extend the usage of a single standard language across an entire national community, first and foremost via universal education, can be considered a precondition for the production of the large quantities of text in standard languages that served as training data for LLMs~\citep{schneider2024sociolinguist}. Educational policies which are rooted in nationalistic and colonial projects~\citep{migge2025material} have further exacerbated this by ensuring that the majority of high-quality digitized texts, such as textbooks, academic papers and documentation, are produced in standardized language. This creates a feedback loop, where LLMs trained on these corpora understand these languages and registers as the means for intellectual discussions and thus generate descriptions in these registers when asked to explain complex and challenging topics. This leads to LLMs considering content that falls outside of these languages and registers to be of lesser intellectual value and less prestigious~\citep{bui-etal-2025-large}. 

Language documentation, which has been framed as an intervention intended to ``save'' endangered languages by capturing their lexico-grammatical structures as data before they are lost~\citep{bird-2020-decolonising} is another example of a language policy initiative with repercussions on LLMs. Language documentation typically involves linguists creating resources, such as orthographies, lexicons, and grammars, to support the production of pedagogical materials for formal language programs, but also increasingly to develop artificial intelligence systems~\citep{himmelmann2006language}. This is often founded on a deep-seated ideological bias of `scriptism'~\citep{scriptism}, which treats written language as the primary, superior, or only ``proper'' form of language, often reducing speech to a mere derivative of text. In the context of language technology and documentation, scriptism manifests as the insistence on standardizing orthographies and prioritizing textual data as a prerequisite for technological support. Scriptism has led LLMs to ignore non-standardised languages by prioritising the creation of massive text-based datasets, which effectively excludes primarily oral or non-standard speech from the ``resource horizon'' of modern AI~\citep{dialectbench}. By treating standardized writing as the only suitable data and by developing systems that disregard linguistic variation, LLM technologies overlook the linguistic practices used by many in the world to communicate~\citep{lutgen2026variation}.

\subsection{Policy Actors and Responses in the LLM Era}

The rapid proliferation of LLMs and Generative AI has led to questions about how language policy can support the development of these technologies in a manner that supports language equality. By analyzing the motivations and limitations of diverse policy actors from Big Tech and state entities to civil society and academic institutions, we argue for a shift towards democratic machine learning strategies that recognize linguistic variation not as statistical noise, but as a vital component of human communication and digital citizenship.

The motivation for big technology companies like Meta, Google or OpenAI to move beyond standardized English is rarely purely altruistic but is governed by a tension between \textit{economic scalability} and \textit{socio-technical responsibility}. Big Tech actors occupy a contradictory space as both the primary enforcers of linguistic standardization and the only entities with the compute power to technically reduce the dialect gap. The dominant logic of the large scale favours standardisation to minimise computational costs; however, increasingly, performance on English cannot easily be improved and, as such, under-resourced and non-standard languages have become strategic for improving overall model performance. Similarly, high-resource language markets, such as English, Spanish or Mandarin, are saturated, while the ``next billion web users''~\citep{arora2019next} speak a diverse range of languages. Development in this logic is likely to remain mostly extractive ~\citep[see also][]{ramponi-2024-language}, unless governed by policies that support linguistic equality. A key technical measure could be the measurement and reporting of a \emph{dialect gap}, which measures the reduction in performance in dialects versus English and by requiring this to be explicitly reported or even directly supported by incentives in government policy. %supported by means of Corporate Social Responsibility (CSR) credits. 
In this manner, the paradigm could be shifted towards more explicit support for non-standardized languages.

State actors have emerged as critical counter-weights to the English-centricity of commercial GenAI, reframing linguistic diversity as a pillar of \textit{digital sovereignty}. Initiatives such as \textbf{OpenEuroLLM}\footnote{\url{https://openeurollm.eu}} support development in particular in EU official languages and the democratic participation of all its citizens. By leveraging public supercomputing infrastructure, such as the \textit{EuroHPC} network\footnote{\url{https://eurohpc.eu}}, states can subsidize the high computational cost of training models on acutely under-resourced dialect data. Furthermore, the implementation of the \textbf{EU AI Act} in 2026 provides a regulatory framework that requires high-risk AI systems to be transparent and non-discriminatory. This provides a legal hook: if an AI used in public administration fails to understand citizens because they use non-standard varieties, it may be deemed a violation of fundamental rights to non-discrimination. While many states now frame linguistic diversity as a public good, they have historically implemented nationalistic policies that marginalize non-standard varieties. This legacy has not only led to the data voids that create issues with current LLMs, but has also reduced the trust among speakers of public initiatives. As such, state-led AI initiatives should avoid replicating the discriminatory practices of the past, e.g. by adopting a policy framework of \textit{linguistic agency}, centered on the principle of ``nothing about us without us'', as similarly suggested by \cite{liu-etal-2022-always}. A democratic strategy~\citep{migge2025material} also requires that speakers of non-standard varieties retain \textit{data sovereignty} over their linguistic repertoires. 

Community and civil society organisations (CSOs) serve as the essential connective tissue between the technical requirements of LLM development and the lived reality of speech communities. For non-standard languages CSOs can transition from being passive subjects of study to active data stewards and algorithmic auditors. Unlike Big Tech's extractive scraping, CSOs can run \emph{citizen science} initiatives~\citep{hilton2021stimmen} (e.g., using platforms like Mozilla Common Voice~\citep{ardila-etal-2020-common}) to collect authentic speech and text with explicit community consent. Furthermore, CSOs can serve as \textit{algorithmic auditors}, performing socio-pragmatic \emph{red-teaming} to identify where models fail to respect local norms or inadvertently enforce standardization. Crucially, these auditing processes must remain reflexive to ensure that community review does not devolve into internal language-policing, wherein specific sub-factions prescriptively gatekeep what constitutes ``authentic'' or ``correct'' regional speech. 

Similarly, academic institutions and national language institutes serve as the primary bridge between technical innovation and socio-historical depth. While Big Tech prioritizes computational scale, universities provide the sociolinguistic granularity necessary to prevent the erasure of non-standard varieties. Furthermore, by fostering interdisciplinary collaboration between Natural Language Processing engineers and sociolinguists, as recently advocated by~\cite{lutgen2026variation}, academics ensure that the models do not merely replicate standardized norms, but reflect a linguistic reality grounded in actual usage.

In the following two sections, we shed light on two case studies, on South Tyrolean German and Kurdish respectively, where we discuss situated implications of language policies in language technology.

\section{Case Study: LLMs and South Tyrolean dialect(s)}\label{sec5}
\label{case_study_tyrolean}
South Tyrol is the northernmost Italian province and is known for its contested history since its annexation to Italy in 1919, its autonomy provisions and its institutional multilingualism. While both German and the Rhaeto-Romance language Ladin were suppressed under Fascism, and speakers of these languages were structurally disadvantaged beyond that, the Autonomy arrangements of 1972 have gradually led to equitable conditions. German is now one of the official languages in the Province and is legally on par with Italian. A tri-partite education system is in place in which either Italian, German or both these languages (plus Ladin as auxiliary language) are used as languages of instruction~\citep[see][for a detailed overview]{platzgummer2021positioning}. However, across large parts of South Tyrol, it is not Standard German, but an ensemble of dialectal varieties assignable to the broader dialect category of Bavarian varieties that are predominantly spoken in social life. Research has stressed the social relevance of these  dialects~\citep{risse2016societal,leonardi2020hardly}, which are also increasingly being used in informal writing, particularly in digital contexts such as on social networks or on WhatsApp (see e.g. \citep{glaznieks2018dialekt,alber2024verschriftungsprinzipien}). Additionally, the relevance of these dialects for constructing belonging has been underlined~\citep{risse2016societal,platzgummer2021positioning,tappeiner2026belonging}. Standard German, in contrast, has been shown to mostly be used as a spoken language in a school context and when interacting with German-speaking tourists~\citep{risse2010zugehorigkeitsstiftendes, leonardi2020hardly}. Most formal writing in South Tyrol still takes place in standard German, with some lexical particularities mainly due to being inserted into a different national context~\citep{abel2018bars}. While South Tyrolean German speakers seem to orient to a standard from Germany as the norm with the highest prestige, they have been shown to consider this standard as somewhat ‘foreign’ to them~\citep{ciccolone2010tutela}.

In the light of this sociolinguistic overview, it stands to reason to expect that speakers of South Tyrolean dialects might benefit from LLMs being able to process their specific varieties. Yet, investigations into South Tyrolean speakers' needs and aspirations for language technology are largely absent. The only exception in this regard is a survey by \cite{blaschke-etal-2024-dialect} studying attitudes towards language technologies of German dialect speakers more broadly. In their analysis, they identify a consistent prioritisation of language technologies working with dialect-input as opposed to dialect-output, as well as of those processing speech rather than written text. These results can easily be linked to sociolinguistic considerations, with the privileging of the spoken over the written mode being analogous to language practices in the respective communities. While their sample consisted of 327 German dialect speakers from seven different European nations, they only checked whether their results differed for the subgroups of respondents from Germany, Austria, and Switzerland. It might be interesting to consider whether speakers' needs and aspirations for language technology are different in sociolinguistic contexts such as South Tyrol, which are characterised by language contact with at least another national language. 

When evaluating the availability of language resources and technologies for South Tyrolean dialects, one first runs up against the problem that South Tyrolean does not have a specific ISO-639 code, which, as previously discussed, means it is not catalogued in NLP resources. As Table~\ref{tab_res_gv} shows, there are only ISO-codes for broader groups of German dialects, such as Upper Saxon, Allemanic, Swabian or Bavarian, which makes only language resources with these codes trackable. South Tyrolean dialects are thereby included within the broader group of Bavarian dialects. While the Table shows that there are some language resources for Bavarian - especially in comparison to many other German dialect groups represented - it is impossible to see at first glance whether this actually also represents South Tyrolean varieties. However, there are some resources available outside of these large collections and a comprehensive list is maintained by \cite{blaschke-etal-2023-survey} on GitHub.\footnote{\url{https://github.com/mainlp/germanic-lrl-corpora}}

\begin{table*}[t]
\centering
\begin{tabular}{p{25mm}ccccc}
\toprule
Name	&ISO Code	&Speakers	&OPUS Words	&VLO Resources\\
\midrule
Upper Saxon	&\texttt{sxu}	&2,000,000	&0	&0\\
Kölsch	&\texttt{ksh}	&250,000	&6,940	&8\\
Palatine	&\texttt{pfl}	&400,000	&122	&0\\
East Franconian	&\texttt{vmf}	&4,900,000	&0	&0\\
Allemanic	&\texttt{gsw}	&7,162,000	&3,030	&1,282\\
Swabian	&\texttt{swg}	&820,000	&203	&8\\
Walser	&\texttt{wae}	&22,780	&0	&0\\
Bavarian	&\texttt{bar}	&15,000,000	&156,237	&124\\
\bottomrule
\end{tabular}
\caption{Representation of German dialects spoken in Germany, Austria,  Switzerland, Liechtenstein or Italy with an assigned ISO code, in major resources for natural language processing. Speaker counts are based on the best information in Wikipedia. OPUS words is the size of the largest resource in \href{https://opus.nlpl.eu/}{OPUS}. VLO refers to the number of listed resources in the \href{https://vlo.clarin.eu/}{CLARIN Virtual Language Obvservatory}.}
\label{tab_res_gv}
\end{table*}

The same then holds for evaluating the availability of pre-trained language models or benchmarks. While, as ~\citet{faisal-etal-2025-dialectal} note, LLMs might work to some degree in dialects - and in fact several Generative AI models responded affirmative when Verena asked \textit{"Versteasch du mi?"} {[Do you understand me? Standard German \textit{Verstehst du mich?}]} - dialects without a specific ISO code are so far not officially supported, and LLMs' performance is not evaluated against them. Moreover, the same is also true for many non-standard varieties with an ISO code. 
%Thus, neither South Tyrolean nor Bavarian appear in any of the relevant multilingual evaluation benchmarks, such as translation-oriented benchmarks like FLORES-200 and NTREX-128, in the cross-lingual topic classification dataset SIB-200~\citep{DBLP:conf/eacl/AdelaniLSVAMGL24}, or in GlobalPIQA~\citep{chang2025globalpiqaevaluatingphysical} or Global MMLU. The only benchmark that Bavarian is included in is the recent \textit{DialectBench}~\cite{dialectbench} mentioned in Section 2.
Consequently, neither South Tyrolean nor Bavarian feature in the massive, general-purpose multilingual evaluation suites commonly used to benchmark commercial LLMs, such as translation-oriented benchmarks (e.g., FLORES-200, NTREX-128), cross-lingual topic classification (SIB-200) \citep{DBLP:conf/eacl/AdelaniLSVAMGL24}, or global reasoning datasets like GlobalPIQA \citep{chang2025globalpiqaevaluatingphysical} and Global MMLU. Despite this absence at the macro-scale data level, there is now a growing ecosystem of specialised dialectal and cross-lingual resources. Both South Tyrolean and regional Bavarian varieties are represented in targeted multilingual tasks, such as the xSID evaluation dataset for slot and intent detection \citep{van-der-goot-etal-2021-masked, winkler-etal-2024-slot} and the multi-dialectal \textit{MaiBaam} treebank within the Universal Dependencies framework \citep{blaschke-etal-2024-maibaam}. Furthermore, Bavarian has been the focus of recent non-multilingual dialectal benchmarks, including the broad-spectrum \textit{DialectBench} \citep{dialectbench} discussed in Section 2, alongside task-specific datasets spanning named entity recognition \citep{peng-etal-2024-sebastian}, automatic speech recognition \citep{Blaschke2025}, information retrieval \citep{litschko-etal-2025-cross} and question answering \citep{winkler-etal-2026-indirect, pei-etal-2026-information}. 
The subsumption of South Tyrolean varieties under the broader category of Bavarian does not necessarily mean that such resources are irrelevant to South Tyrolean: given the linguistic relationship between Bavarian varieties, broader Bavarian resources may provide partial coverage. However, Bavarian encompasses substantial dialectal diversity, as reflected in the mentioned multi-dialectal language resources \citep{blaschke-etal-2024-maibaam,Blaschke2025}, and South Tyrolean varieties additionally have developed within a distinctive multilingual context \citep{risse2016societal,platzgummer2021positioning}. It therefore remains difficult to establish whether improved performance on Bavarian resources translates into adequate representation of South Tyrolean-specific features. More fundamentally, the absence of a South-Tyrolean-specific classification makes it difficult even to measure whether LLMs perform well on South Tyrolean German specifically.

To our knowledge, there is one specific example of the adoption of a \textit{speaker-centric} approach to NLP \citep{ramponi-2024-language}, where a concrete demand from within the community led to an initiative working on fine-tuning models specifically for automated transcription and subtitling of audio(visual) material in South Tyrolean dialects~\citep{ducceschi2025speech}. The use case being addressed is that of transposing the audio into standard German writing, implying that the aim is to allow broader access to non-standard audio(visuals) to speakers of Standard German and/or to provide a basis for machine translating South Tyrolean German into other languages. As \citet{Blaschke2025} have shown by benchmarking several models for Dialect-to-Standard Speech Translation of German dialects from South-East Germany, even the best-performing model exhibited a pronounced gap in performance when comparing transcriptions of dialectal vs. Standard German audio. By fine-tuning Open AI's Speech-to-Text model \textit{Whisper}, \citet{ducceschi2025speech}  have been able to produce higher-quality automatic transcriptions, lowering the Word Error Rate from a baseline of 56\% to 12\%. As the authors note, their model is being applied for large-scale transcription and translation of audiovisual archival material in the context of a heritage collaboration, and its use in news broadcasting and tourism promotion is currently considered. Beyond these use cases, emerging from needs of institutional actors within the community, an improved accuracy of language technologies working with spoken dialect input can be expected to have wider ramifications for both speakers and non-speakers of South Tyrolean German. For the former, this might for instance mean that they might be able to use South Tyrolean varieties with voice-controlled digital assistants such as Siri or Alexa, in digitally mediated interaction that relies on automatic transcripts for record-keeping or other purposes, or in recorded interactions intended also for non-speaker audiences, such as TV broadcasts, or social media content. This last example ties in to the relevance of such technologies for non-speakers, who might benefit from increased accessibility of content in South Tyrolean varieties - which additionally points to the need to include non-speakers or learners into surveys on attitudes towards language technology. Use cases that have not yet been systematically investigated for South Tyrolean varieties - similarly to most other varieties considered as varieties associated with a standard language - are language technologies with dialectal output, neither in written nor in spoken form. This is unfortunate if, like Verena, one would want to avail of raw automatic transcriptions of dialectal audio data into dialectal text for research purposes. Aside from this rather specific use case, comparable patterns across different linguistic contexts suggest that language technologies tend to privilege standard varieties and the linguistic resources associated with them, and rarely represents dialectal variation on its own terms~\citep{ghyselen2020clearing,lonergan2023towards,li2024chinese}.

However, even in the light of new technical developments, it always needs to be kept in mind that, as ~\citet{schneider2022multilingualism} and ~\citet{blaschke-etal-2024-dialect} have shown, whether new capabilities of language technologies actually correspond to user needs and will be taken up in user practices is far from clear - not least because  language technologies are currently ideologically associated first and foremost with English, and to a lesser degree with other standard language varieties, but certainly not with non-standard language varieties. This ideological pattern, as well as the fact that speakers of varieties of South Tyrolean German generally consider themselves proficient enough in Standard German~\citep{stat_report_key}, suggest that speakers will simply make use of Standard German when availing of language technologies. The possibility that this represents an actual risk to the vitality of South Tyrolean German and its community of speakers seems remote at present, though it cannot be excluded. This presents a stark contrast to that of Kurdish varieties, which we turn to in the following section.

\section{Case Study: LLMs and Kurdish Varieties}
\label{case_study_kurdish}

\begin{table*}[t]
\centering
%\small
\begin{tabular}{l c c}
\toprule
\textbf{Variety} &
\textbf{Supported Domains} &
\textbf{Score (/4)} \\
\midrule
Central Kurdish (Sorani, \texttt{ckb})    & Education, Media, Press, Official & 4 \\
Northern Kurdish (Kurmanji, \texttt{kmr})  & Education, Media, Press, Official & 4 \\
Southern Kurdish (\texttt{sdh})           & Press, Publishing & 2 \\
Zazaki (\texttt{zza})                     & Media, Publishing & 2 \\
Hawrami (Gorani, \texttt{hac})          & Publishing & 1 \\
Laki (\texttt{lki})                      & None & 0 \\
\bottomrule
\end{tabular}
\caption{Distribution of institutional support for Kurdish varieties across key sociolinguistic domains (education, media, publishing, and official use). Northern and Central Kurdish shows the widest domain coverage, while other varieties remain restricted to non-institutional or community-based domains.}
\label{table_socio}
\end{table*}

Kurdish is an Indo-European language belonging to the Northwestern Iranian branch, spoken by over 40 million people across Western Asia, primarily in Iraq, Turkey, Iran, Syria, and Armenia, as well as among substantial diaspora communities worldwide~\citep{haig2002kurdish}. Rather than constituting a single unified language, Kurdish represents a dialect continuum comprising several distinct varieties with varying degrees of mutual intelligibility. The principal varieties include Northern Kurdish (Kurmanji), spoken by an estimated 15–20 million speakers predominantly in Turkey, Syria, northern Iraq, and northwestern Iran; Central Kurdish (Sorani), with approximately 10 million speakers concentrated in Iraqi Kurdistan and Iranian Kurdistan; and Southern Kurdish, spoken in Kermanshah, Ilam, and Lorestan~\citep{matras2019revisiting}. Additionally, the Zaza-Gorani languages, including Zazaki and Hawrami, are spoken by communities who identify as ethnic Kurds, though their linguistic classification remains debated; some scholars group them within the broader Kurdish language family while others consider them closely related but distinct Northwestern Iranian languages \citep{arslan2019language}.

\subsection{Language Suppression and Standardization Challenges}

Divided across multiple nation-states, Kurdish has historically faced systematic suppression and assimilation campaigns, including outright bans on public use in Turkey (1923–1991)~\citep{ergil2000kurdish} and Persianization and Arabization policies targeting its use in formal contexts~\cite[p.~268]{romano2025arabization}. \cite{weisi2021language} documents how such policies have led Kurdish-speaking parents in certain regions to use the dominant state language with their children, contributing to intergenerational language shift. Nevertheless, the communicative spaces for Kurdish in media and education have expanded over the past decades, mainly thanks to the establishment of the Kurdistan Regional Government in Iraq, where Kurdish now enjoys official status and institutional support~\citep{opengin2012sociolinguistic}. Unlike Northern Kurdish and Central Kurdish, other varieties such as Southern Kurdish, Hawrami, and Zazaki remain severely under-represented, with Hawrami classified as ``definitely endangered'' by UNESCO \citep{moseley2010atlas}. 

The orthographic landscape of Kurdish reflects its political fragmentation. Northern Kurdish is written using a Latin-based alphabet, while Central Kurdish employs an Arabic-based script. Soviet-era Kurdish communities used a Cyrillic-based system. This orthographic diversity, while enabling written expression within political boundaries, creates significant barriers to cross-dialectal communication, pan-Kurdish linguistic unity \citep{hassanpour2012indivisibility} and efficient language technology~\citep{DBLP:conf/acl/AhmadiA23}.

Table~\ref{table_socio} summarizes the distribution of institutional support across Kurdish varieties in four key sociolinguistic domains: education, media, publishing, and official use. Central Kurdish and Northern Kurdish, the two most widely spoken varieties that are also standardised to some degree~\citep{matras2019revisiting}, exhibit full domain coverage, benefiting from formal recognition in educational curricula, broadcast media, print press, and governmental functions. In contrast, Southern Kurdish and Zazaki maintain a more limited presence, primarily confined to press and publishing activities, reflecting their exclusion from official and educational institutions. Hawrami receives support only in publishing despite ongoing efforts for its official recognition along with Central Kurdish and Northern Kurdish~\citep{sheyholislami2017language, gialdini2023one}. As the least supported variety, Laki lacks institutional backing entirely. This uneven distribution underscores the hierarchical nature of language vitality within the Kurdish continuum, where political recognition and demographic weight correlate strongly with institutional investment, a disparity that poses significant challenges for language preservation efforts and the development of inclusive Kurdish language technologies.

\defcitealias{awlla2025sentiment}{KuBERT}

\begin{table*}[t]
\centering
\resizebox{\textwidth}{!}{%
\begin{tabular}{l|ll|cccc|cc} 
\toprule
\multirow{2}{*}{Variety} & \multicolumn{2}{|c|}{Data} & \multicolumn{4}{c|}{Models} & \multicolumn{2}{c}{Machine Translation} \\  % Seamless
 & Bitext & Audio & XLM-R  & BERT & \texttt{MADLAD-400} & TranslateGemma & Google & Microsoft \\ \hline
 Central Kurdish & $<$300M & $~$200h & \crosstick &  \citetalias{awlla2025sentiment} &  \checkmark & \checkmark & \checkmark & \checkmark \\
Northern Kurdish &  $<$300M & $~$200h & \checkmark & \crosstick & \checkmark & \crosstick & \checkmark & \checkmark \\
Southern Kurdish &  $<$10M & $~$10h & \crosstick & \crosstick & \crosstick & \crosstick  & \crosstick & \crosstick \\
Laki &  $<$1M & $~$2h & \crosstick &  \crosstick & \crosstick & \crosstick & \crosstick & \crosstick \\
Zazaki &  $<$10M & $~$2h  & \crosstick & \crosstick & \crosstick & \crosstick & \crosstick & \crosstick \\
Hawrami &  $<$10M & $~$20h  & \crosstick  & \crosstick& \crosstick & \crosstick & \crosstick & \crosstick \\
\bottomrule
\end{tabular}
}
\caption{Resourcefulness from the perspective of datasets, pretrained models and machine translation}
\label{tab_data_model_mt}
\end{table*}

\begin{table*}[t]
\centering
\begin{tabular}{lccccc} 
\toprule
Variety & FLORES-200 & NTREX-128 & SIB-200 &  GlobalPIQA & Global MMLU \\ \hline
Central Kurdish & \checkmark & \checkmark & \checkmark & \checkmark & \crosstick  \\
Northern Kurdish & \checkmark  & \checkmark & \checkmark & \crosstick &  \crosstick  \\
Southern Kurdish & \crosstick & \crosstick & \crosstick & \crosstick &  \crosstick \\
Laki & \crosstick & \crosstick & \crosstick & \crosstick & \crosstick  \\
Zazaki & \crosstick & \crosstick & \crosstick & \crosstick &  \crosstick  \\
Hawrami & \crosstick & \crosstick & \crosstick & \crosstick & \crosstick  \\
\bottomrule
\end{tabular}
\caption{Resourcefulness from the perspective of benchmarks}
\label{tab_benchmarks}
\end{table*}

\subsection{Implications for Computational Linguistics and NLP}
The sociolinguistic situation of Kurdish has direct implications for its computational processing and representation in language and speech technologies. The historical suppression of the language resulted in limited written corpora, leading to Kurdish being consistently classified as a low-resource language in NLP research. A survey of the NLP literature carried out by Sina reveals that among over 100 papers published in this field, only a handful address varieties other than Central and Northern Kurdish. To further illustrate this disparity, we assess resourcefulness from the three essential pillars of modern language technology: (1) data (text and audio) needed for training models, (2) pretrained models, including embeddings essential for representation, and (3) evaluation benchmarks. Additionally, given its importance as an NLP application, we consider machine translation support as an indicator of technological investment.

Table~\ref{tab_data_model_mt} presents an overview of available resources across Kurdish varieties in terms of parallel text corpora, audio data, pretrained language models, and machine translation services. Central Kurdish followed by Northern Kurdish emerges as the most resourced variety, with over 300 million tokens of textual data and approximately 200 hours of transcribed audio, alongside support from multilingual models such as \texttt{MADLAD-400}~\citep{DBLP:conf/nips/KuduguntaC0GXKS23} and TranslateGemma~\citep{finkelstein2026translategemmatechnicalreport}, as well as commercial translation services from Google and Microsoft. Critically, Southern Kurdish, Laki, Zazaki, and Hawrami remain entirely unsupported across all categories, reflecting a near-total absence from the modern NLP ecosystem.

Table~\ref{tab_benchmarks} provides a complementary perspective by surveying the inclusion of Kurdish varieties in prominent multilingual evaluation benchmarks. Benchmarks are essential for assessing LLM performance in an era of rapid technological advancement. Central and Northern Kurdish appear in translation-oriented benchmarks such as FLORES-200 and NTREX-128, as well as in the cross-lingual topic classification dataset SIB-200~\citep{DBLP:conf/eacl/AdelaniLSVAMGL24}. Central Kurdish is additionally represented in GlobalPIQA~\citep{chang2025globalpiqaevaluatingphysical} for commonsense reasoning. However, neither variety is included in Global MMLU, and the remaining four varieties—Southern Kurdish, Laki, Zazaki, and Hawrami—are absent from all surveyed benchmarks entirely.

It should be noted that the models and benchmarks presented here are not intended to be exhaustive; rather, they are selected to highlight the stark discrepancy in computational support among varieties of Kurdish. This technological marginalization mirrors and reinforces the sociolinguistic hierarchies discussed earlier, posing substantial barriers to the development of inclusive, pan-Kurdish language technologies.

\subsection{The Vicious Circle of Underinvestment and Closed Resources}
The persistent underrepresentation of Kurdish varieties in NLP can be attributed, in part, to a structural lack of investment from governments and policymakers. Unlike languages that benefit from state-sponsored digitization initiatives, corpus development programs, or dedicated research funding, Kurdish, particularly its lesser-resourced varieties, has received negligible institutional support for computational linguistics research. This absence of top-down investment stands in stark contrast to the resources allocated to dominant state languages in the regions where Kurdish is spoken~\citep{DBLP:conf/acl/AhmadiSKMFBHHDA25}.

In the absence of such funding, the development of modern LLMs has come to rely predominantly on publicly available web data, operationalizing the concept of the ``Web as corpus''~\citep{kilgarriff2003introduction}. However, this approach inherently disadvantages languages with limited digital presence, creating a feedback loop in which low online visibility leads to exclusion from training corpora, which in turn perpetuates technological marginalization. For Kurdish, the historical suppression of the language in public and institutional domains has directly constrained its digital footprint, leaving vast gaps in web-crawled datasets.

Moreover, where data collection efforts do exist, whether by individual researchers, diaspora communities, or private enterprises, the resulting resources frequently remain proprietary or unpublished. This reluctance or inability to release data under open-source licenses compounds the scarcity problem, as subsequent research cannot build upon prior work. The result is a vicious circle: the lack of open resources discourages new investment, while the absence of investment limits the creation of shareable datasets. Until this cycle is disrupted through coordinated open-source initiatives, sustained funding, or policy intervention, the majority of Kurdish varieties will remain stranded at the margins of language technology development.

The risk is particularly acute for Southern Kurdish, the persistent underrepresentation of which in language technologies even poses a significant risk of accelerating sociolinguistic bifurcation among its speakers. If NLP developers default to adapting Central Kurdish materials for Southern Kurdish users as a pragmatic technological shortcut, they risk reproducing an intra-linguistic hierarchy. This lack of dedicated digital support is likely to divide Southern Kurdish speakers into two trajectories: one group may shift toward Central Kurdish in digital spaces, drawn by its shared Perso-Arabic script and official institutional status. A second group, however, may abandon Kurdish altogether in digital contexts, shifting instead to the dominant state languages. Since the majority of speakers of Southern Kurdish under 50 are already proficient in the official languages of their respective states (Persian for Iran and Arabic for Iraq), weakened ethnolinguistic vitality in the digital sphere makes assimilation into these dominant languages highly likely. Consequently, while treating Central Kurdish as a baseline for other varieties might bridge a digital gap, it also actively reproduces a standardization hierarchy that threatens the maintenance of Southern Kurdish as a distinct linguistic community.
\section{Conclusion}

This paper has examined the relationship between Large Language Models (LLMs), generative AI, and non-standard language varieties through an interdisciplinary dialogue between sociolinguistics and computational linguistics. Across both perspectives, a common pattern emerges: \emph{contemporary language technologies are deeply shaped by historical processes of linguistic standardization that privileged particular written and codified forms while marginalizing dialects, oral varieties, and less-resourced languages}. Rather than existing outside these sociohistorical dynamics, LLMs are built upon them, drawing primarily on large corpora of standardized written text and thereby reproducing linguistic hierarchies that have their roots in earlier processes of nation-building, literacy expansion, and language standardisation \citep{deumert2010imbodela, schneider2024sociolinguist, migge2025material}.

Our analysis has shown that these inequalities are not simply the result of uneven data availability but are embedded throughout the technical architecture of language technologies. From language identification and tokenization practices to benchmark design and model evaluation, current systems systematically favour standardized, high-resource languages while placing non-standard varieties at a disadvantage \citep{petrov2023language,ahia-etal-2023-languages,xuan2025mmluproxmultilingualbenchmarkadvanced}. However, the pathways through which varieties become marginalized are not uniform. The case of South Tyrolean German demonstrates how varieties can remain largely invisible to language technologies despite their vitality and widespread use, primarily because of issues of scale, categorization, and limited representation within computational infrastructures. The Kurdish case, by contrast, illustrates how technological underrepresentation can intersect with histories of political suppression, uneven institutional recognition, and differential investment across varieties, producing more profound forms of exclusion. Together, these cases highlight the diverse forms that linguistic marginalization can take within contemporary AI systems.

By bringing sociolinguistic and computational linguistic perspectives into direct conversation, this paper has argued that the exclusion of non-standard varieties cannot be understood solely as a technical problem. Rather, it emerges from the interaction of technical infrastructures, evaluation regimes, language ideologies, and broader political and historical processes. Understanding these different configurations of marginalization is essential if we are to move beyond accounts that reduce linguistic inequality in AI to questions of data scarcity alone. In this respect, our analysis echoes recent calls to place language communities and linguistic practices, rather than technological capabilities alone, at the centre of research on language technologies \citep{ramponi-2024-language,bird-2020-decolonising}.

Future work on language technologies should therefore treat linguistic variation as a fundamental characteristic of human communication rather than as noise to be eliminated \citep{lutgen2026variation}. This requires greater attention to the ways in which non-standard varieties are represented in datasets, tokenization schemes, benchmarks, and evaluation practices, as well as more systematic engagement with the needs, aspirations, and communicative realities of speech communities themselves \citep{blaschke-etal-2023-survey, liu-etal-2022-always}. Equally important is the development of evaluation frameworks that are sensitive not only to linguistic performance but also to sociolinguistic context, recognising that varieties differ in their histories, functions, and relationships to standard languages.% \citep{faisal2024dialectbenchnlpbenchmarkdialects, wu2025bitterlessonlearned2000,lutgen2026variation}.

Ultimately, the question of whether LLMs can engage effectively with non-standard varieties is not only about model performance but also about whose linguistic practices become visible within the digital infrastructures that increasingly mediate social, cultural, and civic participation. As generative AI becomes more deeply embedded in everyday life, ensuring that linguistic diversity is meaningfully represented within these systems will be an important component of fostering more equitable and inclusive digital futures.% \citep{kelly2019multilingualism, schneider2022multilingualism,lutgen2026variation}.

%The development of LLMs for non-standard languages represents more than a technical challenge; it is a battleground for linguistic and digital sovereignty, cultural preservation, and social justice. As we have argued, the digital language divide is maintained by a complex interplay of market forces, historical state policies, and the inherent biases of standardized data pipelines. The cases of South Tyrolean and Kurdish varieties, albeit exhibiting major differences, both illustrate how linguistic hierarchies, rooted in standardization, colonialism, and scriptism, are reproduced and intensified through tokenization practices, morphological biases, and evaluation frameworks that privilege standardised, high‑resource languages. Ultimately, closing the dialect gap requires a multi-faceted approach. We advocate for: 

%\begin{description}
    %\item[- \textbf{Big Tech Accountability}] The implementation of dialect gap reporting to incentivize support for under-resourced varieties. 
    %\item[- \textbf{Linguistic Agency}] Adopting the principle of ``nothing about us without us,'' ensuring that communities retain data sovereignty and lead the creation of modern technological terminology. 
    %\item [- \textbf{Interdisciplinary Synthesis}] Leveraging the expertise of academics and language institutes to provide the sociolinguistic granularity that prevents ``hallucinated standardization.'' 
%\end{description}

%By moving away from extractive data practices and toward participatory stewardship, we can ensure that AI serves to safeguard rather than erase large portions of linguistic diversity. 

\section*{Acknowledgements}
The authors would like to thank Verena Blaschke for her feedback on an earlier version of this manuscript, as well as the anonymous reviewers for their constructive comments. Sina Ahmadi gratefully thanks the support of the UZH Grant (reference number 269093).

\begin{appendices}

\end{appendices}

\bibliographystyle{acl_natbib}
\bibliography{references}

\end{document}